\documentclass{llncs}

\usepackage{amssymb}
\usepackage{amsmath}
\usepackage{graphicx}
\graphicspath{{./figures/}}
\usepackage{graphicx}
\usepackage{subfig}
\usepackage{float}
\usepackage{stfloats}
\DeclareMathOperator*{\argmaxC}{\arg\max} 

\usepackage[normalem]{ulem}
\usepackage{color}

\begin{document}
\setlength{\abovedisplayskip}{2pt}
\setlength{\belowdisplayskip}{2pt}
\newcommand{\jd}[1]{\color{blue}#1 \color{black}}
\newcommand{\lw}[1]{\color{red}#1 \color{black}}

% \titlespacing\section{0pt}{12pt plus 4pt minus 2pt}{0pt plus 2pt minus 2pt}
% \titlespacing\subsection{0pt}{12pt plus 4pt minus 2pt}{0pt plus 2pt minus 2pt}
% \titlespacing\subsubsection{0pt}{12pt plus 4pt minus 2pt}{0pt plus 2pt minus 2pt}

\title{High-dimensional Bayesian Optimization of Personalized Cardiac Model Parameters via \\an Embedded Generative Model}
\titlerunning{High-dimensional Bayesian Optimization via an  Embedded VAE}  % abbreviated title (for running head)
\author{Jwala Dhamala\textsuperscript{1} \and Sandesh Ghimire\textsuperscript{1}\and John L. Sapp\textsuperscript{2} \and\\  Milan Horacek\textsuperscript{2} \and Linwei Wang\textsuperscript{1}%
	% index{Dhamala,Jwala} 
	%  index{Wang, Linwei} 
	%   index{Sapp, John} 
	%    index{Horacek,Milan} 
%\thanks{Please note that the LNCS Editorial assumes that all authors have used
%the western naming convention, with given names preceding surnames. This determines
%the structure of the names in the running heads and the author index.}%
}
%\institue{Rochester Institute of Technology, Rochester, NY, 14623, USA}
% (feature abused for this document to repeat the title also on  hand pages)

\institute{\textsuperscript{1} Rochester Institute of Technology, Rochester, NY, 14623, USA\\
	\textsuperscript{2} Dalhousie University, Halifax, Canada
%\textsuperscript{3} Department of Electrical and Computer Engineering, Dalhousie
%	University, Halifax, Canada
%\url{http://www.springer.com/lncs}
}
\authorrunning{Dhamala et al.} % abbreviated author list (for running head)
 \tocauthor{***
 %Ivar Ekeland, Roger Temam, Jeffrey Dean, David Grove,
% Craig Chambers, Kim B. Bruce, and Elisa Bertino
}
%
%\institute{***, ***\\
% \email{I.Ekeland@princeton.edu},\\ WWW home page:
% \texttt{http://users/\homedir iekeland/web/welcome.html}
% \and
% Universit\'{e} de Paris-Sud,
% Laboratoire d'Analyse Num\'{e}rique, B\^{a}timent 425,\\
% F-91405 Orsay Cedex, France
%}

\maketitle              % typeset the title of the contribution

\begin{abstract}
The estimation of patient-specific 
tissue properties in the form of 
model parameters is important for 
personalized physiological models. 
However, these 
tissue properties are spatially varying 
across the underlying  
anatomical model, 
presenting a significance challenge 
of high-dimensional (HD) optimization at the presence of limited measurement data. 
A common solution to reduce the dimension of the parameter space is to explicitly partition the anatomical mesh,  either into a fixed small number of segments 
%for a low-dimensional optimization, 
or a multi-scale hierarchy.  
%for a coarse-to-fine optimization. 
This anatomy-based reduction of 
parameter space 
%explicit division of cardiac mesh
presents a fundamental bottleneck 
to parameter estimation, 
resulting in solutions 
that are either too low in resolution 
to reflect tissue heterogeneity, 
or too high in dimension  
to be reliably estimated within feasible  computation. 
In this paper, we present a novel concept that embeds a generative variational auto-encoder (VAE) into the objective function 
of 
%approach that 
%optimization method that replaces the explicit division of cardiac mesh 
%with an implicit low-dimensional (LD) encoding, achieved by 
%embeds within the 
Bayesian optimization, 
%is able to work with  
%with an embedded stochastic generative model that describes the generation of HD spatially-varying tissue properties from 
%a low-dimensional (LD) manifold. 
%Trained with a variational auto-encoder (VAE), 
%this generative model 
%is embedded within the objective function to provides 
providing an implicit low-dimensional 
(LD) search space that represents the generative code of the HD spatially-varying tissue properties. 
In addition, 
the VAE-encoded knowledge about the generative code is further used to guide the exploration of the search space. 
The presented method is applied to estimating tissue excitability in a cardiac electrophysiological model. 
Synthetic and real-data experiments demonstrate its ability to improve the accuracy of parameter estimation with more than 10x gain in efficiency.

% The abstract should summarize the contents of the paper
% using at least 70 and at most 150 words. It will be set in 9-point
% font size and be inset 1.0 cm from the right and left margins.
% There will be two blank lines before and after the Abstract. \dots
\end{abstract}
\section{Introduction}
Patient-specific simulation models of the heart have shown increasing potential in personalized treatment of cardiac diseases ~\cite{arevalo2016arrhythmia,sermesant2012patient}. %While advances in medical imaging and image analysis have enabled accurate patient-specific anatomical modeling at high resolution, 
The estimation of patient-specific tissue properties, however, remains an unresolved problem. 
One significant challenge is that the unknown tissue properties (in the form of model parameters) are spatially varying at a resolution associated with the discrete cardiac mesh. To estimate these high-dimensional (HD) model parameters is not only algorithmically difficult given indirect and sparse measurements, 
but also computationally intractable in the presence of computing-intensive simulation models. 

Numerous efforts have been made to circumvent the challenge of HD % parameter space %\textit{curse-of-dimensionality}
%in 
parameter estimation. Many works assume homogeneous tissue property throughout the myocardium, which can be represented by a single global model parameter~\cite{giffard2017noninvasive,le2016mri}. To preserve local information about the spatially distributed tissue properties, most existing works resort to dimensionality reduction through an explicit partitioning of the cardiac mesh. These efforts can be generally summarized in two categories. In the first approach, the cardiac mesh is pre-divided into 3-26 segments, each represented by a uniform parameter value~\cite{wong2015velocity,zettinig2013fast}. Naturally, this artificial low-resolution division of the cardiac mesh has a limited ability to represent the underlying tissue heterogeneity that is not known \emph{a priori}. In addition, it has been shown that the initialization of model parameters becomes increasingly more critical as the number of segments grows~\cite{wong2015velocity}. In an alternative approach, the explicit partitioning of the cardiac mesh is done through a coarse-to-fine optimization along a pre-defined multi-scale hierarchy of the cardiac mesh, enabling spatially-adaptive resolution of tissue properties that is higher in certain regions than the others~\cite{chinchapatnam2009estimation,dhamala2016spatially,dhamala2017quantifying,dhamala2017spatially}. However, the representation ability of the final partition is limited by the inflexibility of the pre-defined multi-scale hierarchy: homogeneous regions distributed across different scales cannot be grouped into the same partition, while the resolution of heterogeneous regions can be limited by the level of the scale the optimization can reach \cite{dhamala2016spatially}. In addition, because these methods involve a cascade of optimizations along the coarse-to-fine hierarchy of the cardiac mesh, they are computationally expensive. In the context of parameter estimation for models that could require hours or days for a single simulation, these methods could quickly become computationally prohibitive. %\sout{Hence, an optimal compression of the domain of cardiac mesh is still not attainable}.

In this paper, we present a novel HD parameter optimization approach that replaces the explicitly defined low-dimensional (LD) or multi-scale representation of the parameter space with an implicit LD latent encoding, achieved by embedding within the optimization a stochastic LD-to-HD generative model that describes the generation of the HD spatially-varying tissue properties from a LD manifold. This generative model is obtained with a variational auto-encoder (VAE), trained from a large set of spatially-varying tissue properties reflecting regional tissue abnormality with various locations, sizes, and distributions. 
%obtained by random region growing on a patient-specific cardiac mesh. 
Once trained, the VAE is integrated with a Bayesian optimization (BO)~\cite{brochu2010tutorial} framework in two novel ways. First, the generative model (the VAE decoder) is embedded within the objective function to provide an implicit LD search space for the optimization of HD parameters. Second, the posterior distribution of the LD latent code as learned from the VAE encoder is used as prior knowledge within the BO for an efficient exploration of the LD manifold. To the best of our knowledge, this is the first work that utilizes a probabilistic generative process within an optimization framework for estimating HD patient-specific model parameters.

The presented method is applied to the estimation of local tissue excitability of a cardiac electrophysiological model using non-invasive electrocardiogram (ECG) data. On both synthetic and real data experiments, the presented method is  compared against existing %parameter estimation 
methods based on explicitly-defined LD \cite{wong2015velocity} or multi-scale representation of the parameter space \cite{dhamala2016spatially}. Experiments demonstrate that the presented method can achieve a drastic reduction in computational cost while improving the accuracy of the estimated parameters. 
Beyond the specific model  
considered in this paper, 
the presented method provides an efficient and reliable solution to a wider range of HD model parameter estimation problems.

\section{Background: Cardiac Electrophysiological System}
\textit{\textbf{Cardiac Electrophysiology Model:}}
Among the different types of cardiac electrophysiological models, phenomenological models such as the Aliev-Panfilov (AP) model~\cite{aliev1996simple} can explain the  macroscopic process of cardiac excitation with a small number of model parameters and reasonable computation, making their use in parameter estimation widespread~\cite{sermesant2012patient,giffard2017noninvasive,dhamala2016spatially}. 
Therefore, in this study, the AP model given below is chosen to test the feasibility of the presented method:
\begin{align}
\begin{split}
		{\partial u}/{\partial t} & = \nabla(\mathbf{D} \nabla u) - cu(u-\theta)(u-1) - uv,\\
		{\partial v}/{\partial t} & = 	\varepsilon(u,v)(-v - cu(u - \theta - 1)),
	\label{eq:AP}
    \end{split}
\end{align}
Here, parameter $\varepsilon$ controls the coupling between the normalized transmembrane action potential $u$ and the recovery current $v$, $\mathbf{D}$ is the diffusion tensor, $\mathit{c}$ controls the repolarization, and $\theta$ controls the excitability of the cell. The transmural action potential is computed by solving the AP model~(\ref{eq:AP}) on a 3D myocardium discretized using the meshfree method~\cite{wang2010physiological}. Here, the output $u$ is most sensitive to the value of the parameter $\theta$ \cite{dhamala2016spatially}, which is associated to the ischemic severity of the myocardial tissue. Therefore, as an initial study, we focus on the estimation of the spatially distributed parameter $\theta$.\\
\textit{\textbf{Body-surface ECG Model:}}
The propagation of the spatio-temporal transmural action potential $\textbf{U}$ to the potentials measured on the body surface $\textbf{Y}$ can be described by the quasi-static approximation of the electromagnetic theory~\cite{plonsey1969bioelectric}. Solving the governing equations on a discrete heart-torso mesh, a linear relationship between $\textbf{U}$ and $\textbf{Y}$ can be obtained as: $\textbf{Y} = \textbf{H}(\textbf{U}(\pmb{\theta}))$, where %$\textbf{H}$ is the transfer matrix that is unique to a heart-torso geometry, and 
$\pmb{\theta}$  is the vector of local parameters $\theta$ at the resolution of the cardiac mesh.   

\section{HD Parameter Estimation via an Embedded Generative Model}
%The three dimensionally distributed parameters
The parameter 
$\pmb{\theta}$ in the AP model~(\ref{eq:AP}) 
can estimated by maximizing the similarity between the measured ECG signal $\textbf{Y}$ and those simulated by the combined cardiac electrophysiological and surface ECG model $M(\pmb{\theta})$: % as follows:
\begin{align}
\hat{\pmb{\theta}} = \argmaxC_{\pmb{\theta}}{-||\textbf{Y} - M(\pmb{\theta})||^2}
\label{eq:objfunc}
\end{align}

\begin{figure*}[!t]
	\centering
    \subfloat{\includegraphics[width=0.65\textwidth]{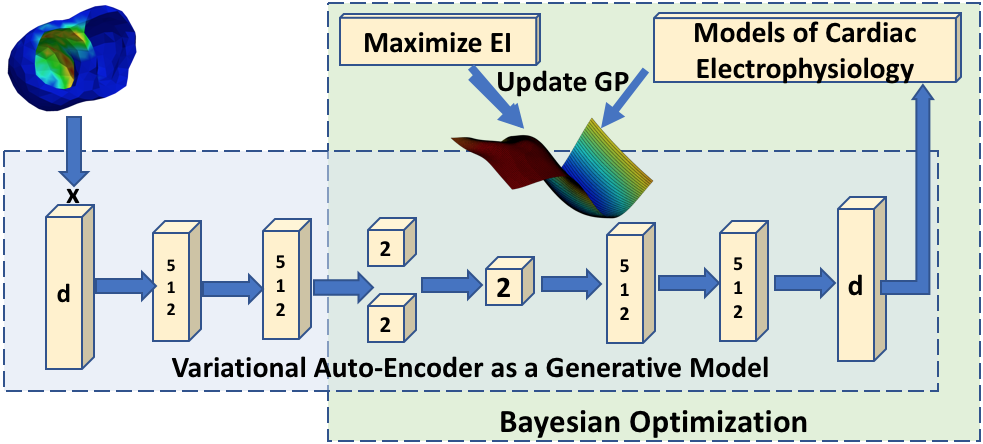}}	 
    \caption{\small{The workflow diagram of the presented high dimensional Bayesian optimization via embedded variational auto-encoder for local parameter estimation.}}
    \label{fig:workflow}
\end{figure*}

The direct estimation of $\pmb{\theta}$ at the resolution of the cardiac mesh is infeasible, both due to limited identifiability from limited ECG data and prohibitively high computational cost. The presented method enables this HD optimization by embedding within the Bayesian optimization framework a stochastic generative model that generates the HD parameter on the cardiac mesh from a LD manifold. As outlined in Fig.~\ref{fig:workflow}, the presented method includes two major components: 1) the construction of a generative model of HD parameters representing spatially-varying tissue properties at the resolution of the cardiac mesh, and 2) a novel Bayesian optimization method utilizing the embedded generative model.

\subsection{LD-to-HD Parameter Generation via VAE}
Recently, generative models have shown promising potential in unsupervised learning of abstract LD representations from which complex images can be generated~\cite{kingma2013auto}. Inspired by this, here, we utilize a VAE to obtain a stochastic generative model of the HD spatially varying tissue properties from a LD manifold.

\textbf{\textit{Generative VAE model:}} We assume that the spatially varying tissue properties at the resolution of a cardiac mesh $\pmb{\theta}$ is generated by a small number of unobserved continuous random variables $\textbf{z}$ in a LD manifold. To obtain the generative process from $\textbf{z}$ to $\pmb{\theta}$, the VAE consists of two modules: a probabilistic deep encoder network with parameters $\pmb{\alpha}$ that approximates the intractable true posterior density $p_{\pmb{\beta}}(\textbf{z}|\pmb{\theta})$ as $q_{\pmb{\alpha}}(\textbf{z}|\pmb{\theta})$; and a probabilistic deep decoder network with parameters $\pmb{\beta}$ that can probabilistically reconstruct $\pmb{\theta}$ given $\textbf{z}$ as $p_{\pmb{\beta}}(\pmb{\theta}|\textbf{z})$. Both 
%the encoder and decoder 
networks consist of three fully-connected layers as shown in Fig.~\ref{fig:workflow}. 

To train the VAE, we generate %training data 
$\pmb{\Theta}=\big\{\pmb{\theta}^{(i)}\big\}_{i=1}^{N}$ consisting of $N$ configurations of heterogeneous tissue properties in a patient-specific cardiac mesh. 
The training involves optimizing the variational lower bound on the marginal likelihood of each training data $\pmb{\theta}^{(i)}$ with respect to network parameters $\pmb{\alpha}$ and $\pmb{\beta}$: %, as given by: 
\begin{align}
\mathcal{L}(\pmb{\alpha};\pmb{\beta};\pmb{\theta}^{(i)}) = -D_{\mathrm{KL}} (q_{\pmb{\alpha}}(\textbf{z}|\pmb{\theta}^{(i)}) || p_{\pmb{\beta}}(\textbf{z})) + E_{q_{\alpha}(\textbf{z}|\pmb{\theta}^{(i)})} [\mathrm{log} p_{\pmb{\beta}}(\pmb{\theta}^{(i)}|\textbf{z})].
\label{eq:loss}
\end{align}
% Here, the first term expected negative reconstruction error between the original and reconstructed tissue properties and . 
where 
we model $p_{\pmb{\beta}}(\pmb{\theta}|\textbf{z})$ with a Bernoulli distribution.
To optimize %the above loss function 
Eq.~(\ref{eq:loss}), stochastic gradient descent with standard backpropagation is utilized. 
%In specific, 
Assuming that the approximate posterior $q_{\alpha}(\textbf{z}|\pmb{\theta})$ is a Gaussian density and the prior $p_{\pmb{\beta}}(\textbf{z})\sim\mathcal{N}(0,1)$, their KL divergence 
%of  $q_{\alpha}(\textbf{z}|\pmb{\theta})$ from $p_{\pmb{\beta}}(\textbf{z})$ 
can be derived analytically as:
\begin{align}
D_{\mathrm{KL}} (q_{\pmb{\alpha}}(\textbf{z}|\pmb{\theta}^{(i)}) || p_{\pmb{\beta}}(\textbf{z})) = -\frac{1}{2} \sum_{j=1}^{N_z}(1+\mathrm{log}(\pmb{\sigma}_j^2)-\pmb{\mu}_j^2 - \pmb{\sigma}_j^2),
\label{eq:kl}
\end{align}
where $N_z$ is the dimension of $\textbf{z}$, and $\pmb{\mu}$ and $\pmb{\sigma}$ are mean and variance from $q_{\pmb{\alpha}}(\textbf{z}|\pmb{\theta}^{(i)})$. Because stochastic latent variables are utilized, the gradient of the expected negative reconstruction term during backpropagation cannot be directly obtained. The popular re-parameterization trick is utilized to express $\textbf{z}$ as a deterministic variable as $\textbf{z}^{(i)} = \pmb{\mu}^{(i)} + \pmb{\sigma}^{(i)} \pmb{\epsilon}$, where $\pmb{\epsilon}\sim\mathcal{N}(0,\textbf{I})$ is noise~\cite{rezende2014stochastic,kingma2013auto}. %

\textbf{\emph{Probabilistic modeling of the latent code:}} After the training of VAE, in addition to the generative model provided by the decoder, 
the encoder provides an approximated conditional density of the LD latent code $q_{\alpha}(\textbf{z}|\pmb{\theta})$. This represents valuable knowledge 
about the probabilistic distribution of $\textbf{z}$ learned from a large number of training data $\pmb{\Theta}$. To utilize this knowledge in the subsequent optimization, we 
integrate $q_{\alpha}(\textbf{z}|\pmb{\theta})$ over the training data  $\pmb{\Theta}$ to obtain the density $q_{\pmb{\alpha}}(\textbf{z})$ as a mixture of Gaussians $1/N\sum_{i}^{N}{\mathcal{N}(\pmb{\mu}^{(i)},\pmb{\Sigma}^{(i)}})$, where $\pmb{\mu}^{(i)}$ and $\pmb{\Sigma}^{(i)}$ are mean and variance from $q_{\pmb{\alpha}}(\textbf{z}|\pmb{\theta}^{(i)})$.
Because the number of mixture components in $q_{\pmb{\alpha}}(\textbf{z})$ scales linearly with the number of training data, we obtain an approximation to $q_{\pmb{\alpha}}(\textbf{z})$ by reducing the number of mixture components in two ways. First, we assume that $q_{\pmb{\alpha}}(\textbf{z})$ can be represented with a single Gaussian density whose mean and variances is calculated as $~\mathcal{N}\big(1/N \sum_{i}^N\pmb{\mu}^{(i)}, 1/N \sum_{i}^N(\pmb{\Sigma}^{(i)} + \pmb{\mu}^{(i)}\pmb{\mu}^{(i)T}) - \pmb{\mu}\pmb{\mu}^{T}\big)$. 
Alternatively, we assume that $q_{\pmb{\alpha}}(\textbf{z})$  can be represented by a mixture of Gaussians with $K<<N$ components. To reduce the number of mixture components from $N$ to $K$, we use k-means clustering with the Bregman divergence~\cite{garcia2009levels} as a similarity metric on the $N$-component Gaussian densities. 

In this way, we obtain a generative model $p_{\pmb{\beta}}(\pmb{\theta}|\textbf{z})$ of HD %spatially-varying
tissue properties from an implicit LD manifold, and prior knowledge of the LD manifold $q_{\pmb{\alpha}}(\textbf{z})$ from the probabilistic encoder. %$q_{\pmb{\alpha}}(\textbf{z}|\pmb{\theta})$. 
Both will be embedded into %the framework of 
Bayesian optimization to enable efficient and accurate HD parameter estimation.

\subsection{Bayesian Optimization with Embedded Generative Model}
Representing the HD parameter $\pmb{\theta}$ 
with the expectation of the trained decoder $p_{\pmb{\beta}}(\pmb{\theta}|\textbf{z})$, 
we embed the generative model 
into a revised objective function: 
\begin{align}
\hat{\textbf{z}} = \argmaxC_{\textbf{z}}{-||\textbf{Y} - M\big(\mathrm{E}[p_{\pmb{\beta}}(\pmb{\theta}|\textbf{z}]\big)||^2}
\label{eq:objfunc_new}
\end{align}
which allow us to optimize the HD parameter $\pmb{\theta}$ in an implicit LD manifold of $\textbf{z}$. 
For Bayesian optimization, we assume a Gaussian process (GP) $\sim\mathcal{N}(\mu(.),\sigma(.,.))$ -- 
with a zero mean function and an anisotropic M\'{a}tern 5/2 co-variance function~\cite{brochu2010tutorial} --  
as a prior over the objective function (~\ref{eq:objfunc_new}). Then, the optimization consists of two iterative steps: first, by maximizing the acquisition function points in the LD manifold that 
allow the GP to both globally approximate Eq. (\ref{eq:objfunc_new}) (exploration) and locally refine the regions of optimum (exploitation) is selected; second, the GP is re-trained with the recently selected point.

\textbf{\textit{Acquisition function informed by VAE-encoded knowledge:}} 
To find optimal points in the LD manifold for updating the GP,
we adopt the expected improvement (EI) function that selects point with maximum expected improvement over the current best objective function value~\cite{brochu2010tutorial}. For a GP posterior, it can be obtained analytically as:
\begin{align}
\mathrm{EI(}\textbf{z})& = (\mu(\textbf{z})-f^+)\mathrm{\Phi}\bigg(\frac{\mu(\textbf{z})-f^\textrm{+}}{\sigma(\textbf{z})}\bigg) + \sigma(\textbf{z}){\phi}\bigg(\frac{\mu(\textbf{z})-f^\textrm{+}}{\sigma(\textbf{z})}\bigg),
\label{eq:ei}
\end{align}
where $\Phi$ is the normal cumulative distribution, $\phi$ is the normal density function, and $f^{+}$ is the maximum of the objective function obtained so far. The first term here controls the exploitation (through high $\mu(\textbf{z})$) and the second term controls exploration (through high $\sigma(\textbf{z})$). Because using only $f^+$ can lead to excessive local exploitation, a common practice is to augment  $f^+$ with a constant trade-off parameter $\varepsilon$ as: $f^{+}+\varepsilon$~\cite{brochu2010tutorial}.
%Commonly, a constant or zero value of $\varepsilon$ is used for an overall increased exploration. 
% %Alternatively, 
% a higher value is used in the 
% earlier iterations
%  of the optimization, 
% which is decreased to allow   
% a decreasing amount of exploration 
% as the optimization continues. 
Here, we utilize the VAE-encoded knowledge about the LD manifold $q_{\pmb{\alpha}}(\textbf{z})$ to enforce higher exploration in the regions of high probability density for $\textbf{z}$, and lower elsewhere. In specific, we define $\varepsilon(\textbf{z}) = -f^{+} \sum_{i=1}^{K}w_i(\textbf{z}-\pmb{\mu}_i) \pmb{\Sigma}_{i}^{-1} (\textbf{z}-\pmb{\mu}_i)$, where $w$, $\pmb{\mu}$ and $\pmb{\Sigma}$ are the weight, mean and variance of the Gaussian mixture components in $q_{\pmb{\alpha}}(\textbf{z})$. 

\textbf{\emph{GP Update:}}
Once a new point $\textbf{z}^{(n)}$ is selected by maximizing EI, the objective function (\ref{eq:objfunc})
is evaluated at the HD parameter given by the mean of $p_{\pmb{\beta}}(\pmb{\theta}|\textbf{z}^{(n)})$ obtained from the embedded generative model. The GP is then re-fitted to the updated data obtained by adding the new pair of $\textbf{z}^{(n)}$ and the corresponding value of the objective function by maximizing its log marginal likelihood with respect to kernel parameters: length scales and covariance amplitude. 

\section{Experiments}
%*******
\begin{figure*}[!t]
	\centering
    \subfloat{\includegraphics[width=1\textwidth]{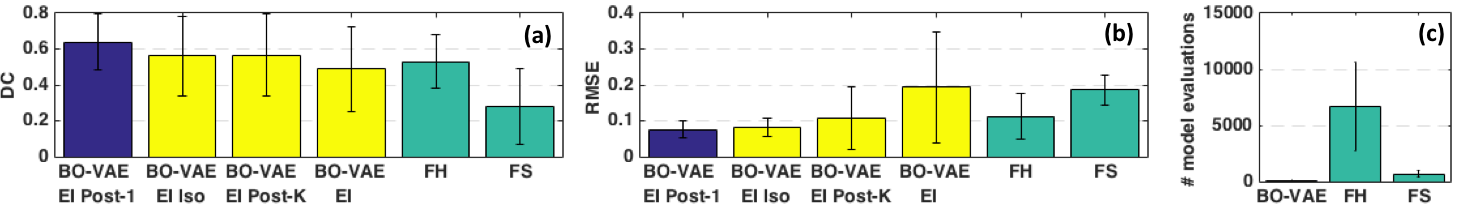}}	    
     \caption{\small{Comparison of  BO-VAE EI post-1 (blue bar) with: 1) FH and FS (green bars); and 2) BO-VAE using standard EI, EI prior, and EI post-K (yellow bars) in terms of DC, RMSE, and number of model evaluations (from left to right).}}
    \label{fig:syn_stat}
\end{figure*}
%*******

\paragraph{\textbf{Synthetic Experiments:}}
We include %a total of 
27 synthetic experiments on three CT-derived human heart-torso models. % constructed from CT-images. 
In each case, an infarct sized $2\%-40\%$ of the heart was placed at differing locations using various combination of the AHA segments. The value of the parameter $\theta$ in the infarcted region and the healthy region is set to $0.5$ and $0.15$, respectively. 120-lead ECG is simulated and corrupted with 20dB Gaussian noise as measurement data. To evaluate the accuracy in estimated parameters with two metrics: 1) root mean square error (RMSE) between the true and estimated parameters; and 2) dice coefficient (DC) = $\frac{2(|S_1 \cap S_2|)}{|S_1| + |S_2|}$,  where $S_1$ and $S_2$ are the sets of nodes in the true and estimated regions of infarct which is determined by using a threshold that minimizes the intra-region variance on the estimated parameter values~\cite{otsu1975threshold}.

\textit{VAE architecture and training:}
For each heart, we generate a training dataset $\pmb{\Theta}=\big\{\pmb{\theta}^{(i)}\big\}_{i=1}^{N}$ corresponding to tissue properties with various heterogeneous infarcts. Each infarct is generated by random region growing in which, starting with one infarct node, one out of the five closest neighbors of the present set of infarct nodes is randomly added as an infarct node. This is repeated until an infarct of desired size is attained which is then added to the training dataset. Because infarcts generated in this fashion tend to be irregular in shape, we also add to the training data infarcts generated by growing the infarct using the closest neighbor to the infarct center. In total, we extracted 123,896, 155,099, and 116,459 training data for each heart. On these dataset, VAE with an architecture as shown in Fig.~\ref{fig:workflow} that consists of an encoder and a decoder network with two hidden layers of 512 hidden units each, a pair of two-dimensional units for the mean and log-variance of the latent code, and sigmoid activation function is trained using \textit{Adam} optimizer. The training time for each dataset using Titan X Maxwell GPU was 9.77, 13.96, and 9.00 minutes, respectively.

%%Intel i7-5820K CPU 3.30GHz with  which is lower than time required for a single model evaluation (3sec- 20mins). %Intel(R) Core(TM) i7-5820K CPU 3.30GHz with Titan X Maxwell GPU.  
\textit{Comparison with existing methods:}
The presented method (termed as BO-VAE) is compared against two common approaches based on explicitly defined LD representation of the cardiac mesh: 1) pre-fixed 18 segments (termed as fixed-segment (FS) method); and  2) partitions along a fixed multi-scale representation of the cardiac mesh obtained during coarse-to-fine optimization (termed as fixed-hierarchy (FH) method). Fig.~\ref{fig:syn_stat}(a)-(b) summarizes the accuracy% of these methods
, demonstrating that BO-VAE (blue bar) is more accurate than the other two methods (green bars) in both DC and RMSE (paired t-tests, p$<0.012$). Fig.~\ref{fig:syn_stat}(d) shows that this improvement in accuracy is achieved at a reduction of computation cost by $\sim87.57\%$ for the FS method and $\sim98.73\%$ for the FH method.
%*******
\begin{figure*}[!t]
	\centering
    \subfloat{\includegraphics[width=0.85\textwidth]{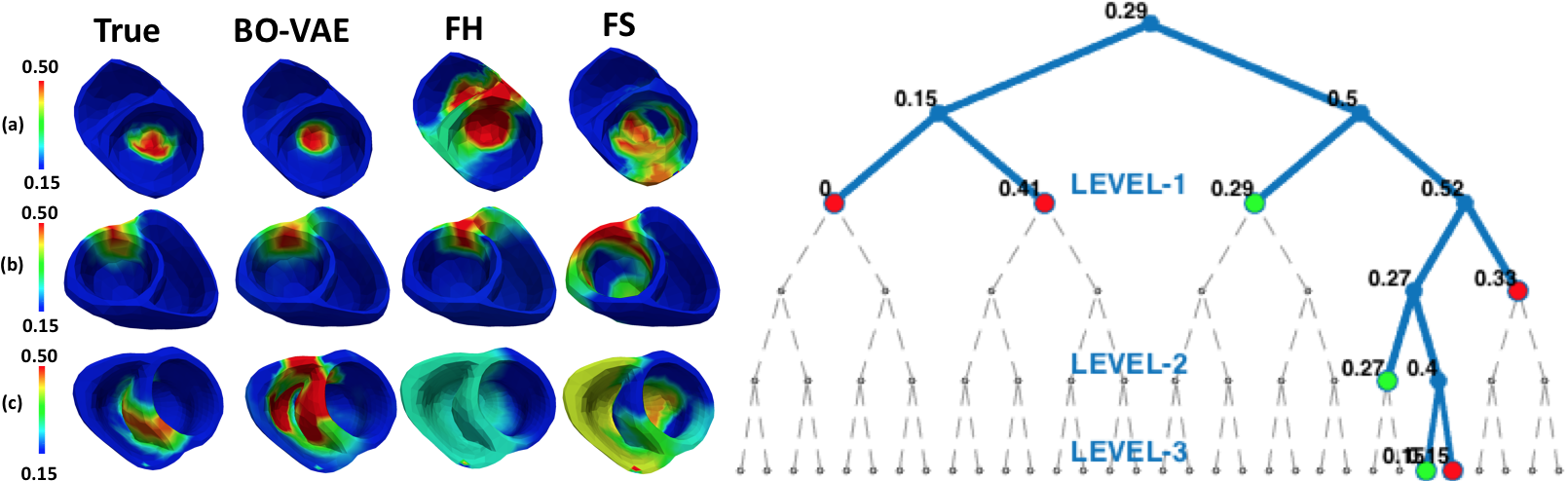}}%	    
    \caption{\small{Left: Examples of estimated parameters with BO-VAE, FH, and FS. Right: Progression of FH on the multi-scale hierarchy for parameter estimation of case (a) (green leaf: homogeneous tissues and red leaf: heterogeneous tissues).}}
    \label{fig:tree}
\end{figure*}
%*****************

\begin{figure*}[!t]
	\centering
    \subfloat{\includegraphics[width=0.85\textwidth]{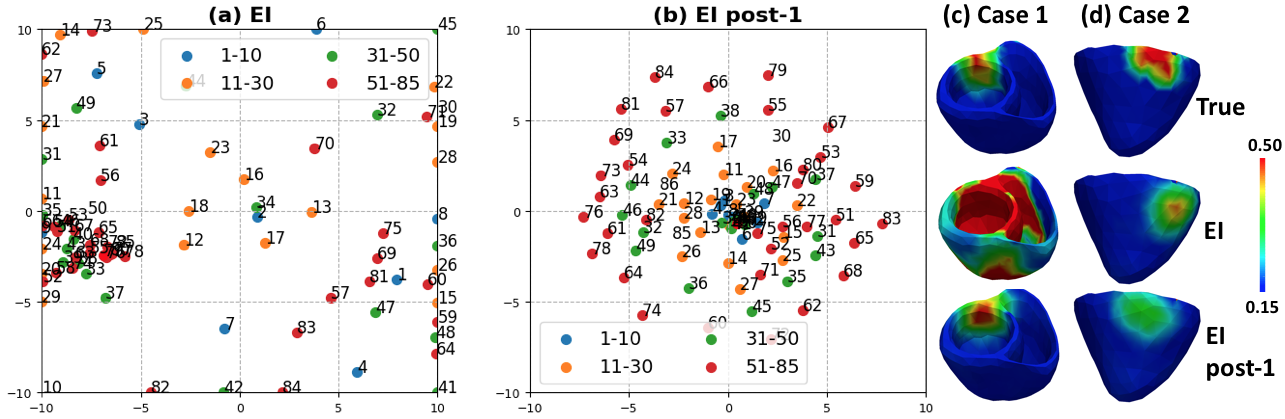}}%
    \caption{\small{Comparison of points selected by EI and EI post-1 shows that with EI post-1 the regions of higher $q_{\alpha}(\textbf{z})$ is explored before the regions of lower $q_{\alpha}(\textbf{z})$.}}
    \label{fig:bo_ei}
\end{figure*}

As expected, the FS method shows the lowest accuracy in which, as illustrated in Fig.~\ref{fig:syn_stat}, the estimated parameters either completely miss the infarct or are associated with a large region of false positives. This could partly be because direct optimization of 18 unknown model parameters without good initialization is difficult, while many of these dimensions are wasted at representing a region of homogeneous tissue. The FH method overcomes this issue to some extent, although it suffers from limited accuracy in many cases primarily due to an inflexible multi-scale hierarchy. An example is shown in the right panel of Fig.~\ref{fig:tree}: several dimensions are wasted at representing homogeneous healthy regions (green nodes) distributed across different scales, which increasingly limits the ability of the optimization to go deeper along the tree to represent heterogeneous tissues at the necessary scale. In contrast, BO-VAE is not limited by such explicitly imposed structure of the parameter space, which allows it to attain higher accuracy with only 2 latent dimensions and a $0.12$ and $0.013$ fraction of the computation time of the FS and FH methods, respectively.\\
\textit{The effect of VAE-encoded knowledge about the LD manifold:}
To study the effect of incorporating the VAE-encoded knowledge of the LD manifold in the EI acquisition function, we compare the standard EI with EI augmented with three types of distributions on $\textbf{z}$: 1) $p_{\pmb{\beta}}(\textbf{z})\sim\mathcal{N}(0,1)$ (EI isotropic), 2) approximated $q_{\pmb{\alpha}}(\textbf{z})$  with a single Gaussian density (EI Post-1); and 3) approximated $q_{\pmb{\alpha}}(\textbf{z})$ with a mixture of Gaussian with 10 components (EI Post-K). As shown in Fig.~\ref{fig:syn_stat}, the estimation accuracy using all three distributions is higher than that without using any knowledge about $\textbf{z}$. In particular, the estimation accuracy with EI Post-1 is the highest. Fig.~\ref{fig:bo_ei} illustrates how the knowledge from $q_{\pmb{\alpha}}(\textbf{z})$ enables a more efficient search of the latent manifold. As shown, when $q_{\pmb{\alpha}}(\textbf{z})$ is utilized, the exploration gradually proceeds from the region of high probability density to the region of low probability density (Fig.~\ref{fig:bo_ei}(b)). In comparison, without anything to guide the placement of the points, they are spread in an attempt to reduce overall variance (Fig.~\ref{fig:bo_ei}(a)). As a result, it could result in incorrect or suboptimal solution as shown in Fig.~\ref{fig:bo_ei} (c) and (d), respectively.

\begin{figure*}[!t]
	\centering   \subfloat{\includegraphics[width=0.85\textwidth]{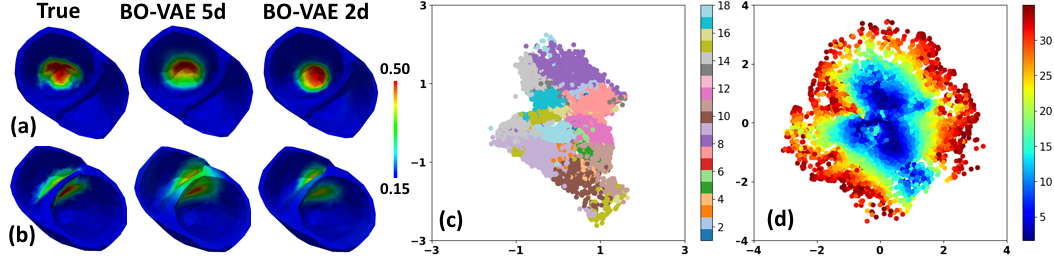}}
    \caption{\small{LD manifold based on: (a) infarct size, and (b) infarct location shows that these information are encoded in the latent code. (c)(d) Examples in which estimated parameters with 5d latent code is more accurate than that with 2d latent code.}}
    \label{fig:latent_code}
\end{figure*}

We also experimented with utilizing a higher-dimensional latent code $\textbf{z}$. Fig.~\ref{fig:latent_code}(c)(d) give examples of the estimated parameters using a five-dimensional (5d) \emph{vs}.\ two-dimensional (2d) latent code, where only a marginal improvement of accuracy is observed with the increase in dimensions of $\textbf{z}$. This could be because, given the focus of the training data on tissue properties resulting from local infarcts, a 2d latent code is sufficient to account for the necessary latent generative factors. Fig.~\ref{fig:latent_code}(a)-(b) shows the plot of these 2d latent codes corresponding to infarcts of different sizes and locations. Interestingly, it appears that the radial direction in the LD manifold accounts for the infarct size (a), while latent codes cluster by infarct location in the LD manifold (b).

%*******
%*******
%*******
\begin{figure*}[!t]
	\centering
    \subfloat{\includegraphics[width=0.7\textwidth]{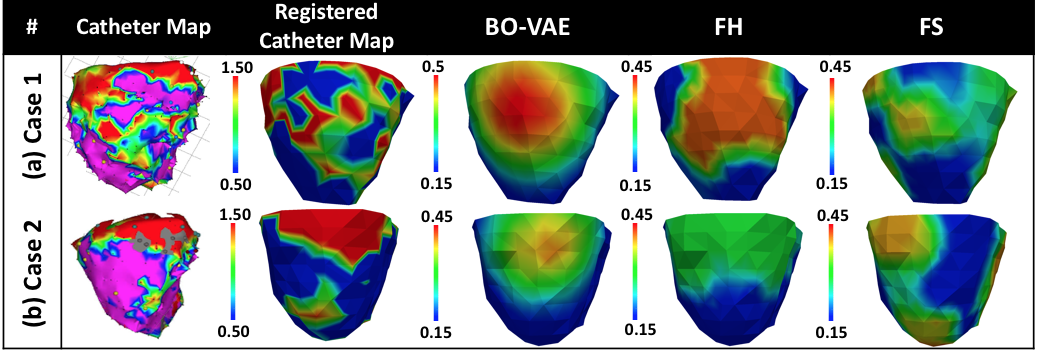}}%	    
    \caption{\small{Model parameter estimated with BO-VAE, FH, and FS on real-data study.}}
    \label{fig:real_fig}
\end{figure*}

%*******
\paragraph{\textbf{Real Data Experiments:}}
Real-data studies are conducted on two patients who underwent catheter ablation of ventricular tachycardia due to previous myocardial infraction. The patient-specific geometrical models of heart and torso are constructed from axial CT images. Using 120-lead ECG as measurement data, we evaluate the performance of the presented method in estimating local parameters in comparison with the FH and FS methods. The accuracy of the estimated parameters is evaluated using the bipolar voltage data from \emph{\textit{in-vivo}} catheter mapping. Note that since the voltage maps are not a direct measure of tissue properties, it is used as a reference rather than ground truth. The first two columns of Fig.~\ref{fig:real_fig} show the reference catheter mapping data (red: dense scar $\le$0.5mV, purple: healthy tissue $>1.5$mV, and green: scar border $0.5-1.5$mV) and the same data registered to CT-derived cardiac meshes.\\ 
% ****
% ****
%\textit{Case 1:}  
%\textit{Case 2:} 
The catheter map in case 1 (Fig.~\ref{fig:real_fig}(a)) shows a highly heterogeneous infarct spread over a large region in the lateral LV region. The estimated parameters by all three methods capture this region of infarct. To attain this accuracy, the FH and FS methods required 4056 and 1058 model evaluations, whereas BO-VAE required only 105 model evaluations. By contrast, as shown in Fig.~\ref{fig:real_fig}(b), case 2 has a smaller region of dense scar in the lateral LV. The estimated model parameters by BO-VAE and FH correctly reveal this region of abnormal tissue, whereas although FS reveals a scar it has less accurate overlap with the low voltage region shown by the voltage map. To attain this level of accuracy, the presented method required only 105 model evaluations in comparison to the FH and FS methods that required 5798 and 1501 model evaluations, respectively.

\section{Conclusion}
We presented a novel approach for estimating HD cardiac model parameters by embedding within the Bayesian optimization framework a generative model of the HD tissue properties 
from an implicit LD manifold. Experiments show a gain in accuracy with drastically reduced computational cost. Future works include two direction: 1) to incorporate more  realistic training data from high resolution 3D imaging for a more expressive generative model and potentially an improved accuracy in estimating highly heterogeneous tissues; and 2) to improve the efficiency by investigating novel ways to incorporate the knowledge of latent manifold to guide the active selection of points during BO.% Bayesian optimization. 

\bibliographystyle{splncs03}
\bibliography{main.bib}

\end{document}